\def\year{2019}\relax
\documentclass[letterpaper]{article} 
\usepackage{aaai19}  
\usepackage{times}  
\usepackage{helvet} 
\usepackage{courier}  
\usepackage[hyphens]{url}  
\usepackage{graphicx} 
\def\UrlFont{\rm}  
\usepackage{graphicx}  
\usepackage{multirow}
\usepackage{amsmath}
\DeclareMathOperator*{\argmax}{arg\,max}

\title{Abductive Reasoning as Self-Supervision for Common Sense Question Answering}
\author{Sathyanarayanan N. Aakur\textsuperscript{\rm 1,2}, Sudeep Sarkar\textsuperscript{\rm 1} \\
\textsuperscript{\rm 1}Oklahoma State University, \textsuperscript{\rm 2}University of South Florida}
 
\begin{document}

\maketitle

\begin{abstract}
Question answering has seen significant advances in recent times, especially with the introduction of increasingly bigger transformer-based models pre-trained on massive amounts of data. 
While achieving impressive results on many benchmarks, their performances appear to be proportional to the amount of training data available in the target domain. 
In this work, we explore the ability of current question-answering models to generalize - to both other domains as well as with restricted training data. 
We find that large amounts of training data are necessary, both for pre-training as well as fine-tuning to a task, for the models to perform well on the designated task. 
We introduce a novel abductive reasoning approach based on Grenander's Pattern Theory framework to provide self-supervised domain adaptation cues or "pseudo-labels," which can be used instead of expensive human annotations. 
The proposed self-supervised training regimen allows for effective domain adaptation without losing performance compared to fully supervised baselines. 
Extensive experiments on two publicly available benchmarks show the efficacy of the proposed approach. 
We show that neural networks models trained using self-labeled data can retain up to $75\%$ of the performance of models trained on large amounts of human-annotated training data. 
\end{abstract}

\section{Introduction}
Language models such as BERT~\cite{devlin2018bert} and GPT~\cite{radford2019language} have shown remarkable progress in many natural language processing tasks through vigorous pre-training for language models. They have achieved state-of-the-art performance on several natural language inference (NLI) task~\cite{wang2018glue}, even surpassing human-level performance on some benchmarks. Despite such great success, it appears that the underlying task, that of \textit{commonsense natural language inference} is not yet solved. Unfortunately, there does seem to exist a strong correlation between the quantity and quality of training data available to these models and their ability to achieve high accuracy. Given the dependency on large amounts of \emph{expensive}, \emph{annotated} data, the ability of such models to generalize to a new domain, or even the same domain with adversarial artifacts remains limited. The absence of ``\emph{common sense knowledge}'' such as that about the world, concepts and semantic relationships prevent the models demonstrating complete \emph{understanding} their world and hence behaving reasonably in unforeseen situations~\cite{gunning2018machine}. 

\begin{figure}[h]
\centering
\includegraphics[width=0.99\columnwidth]{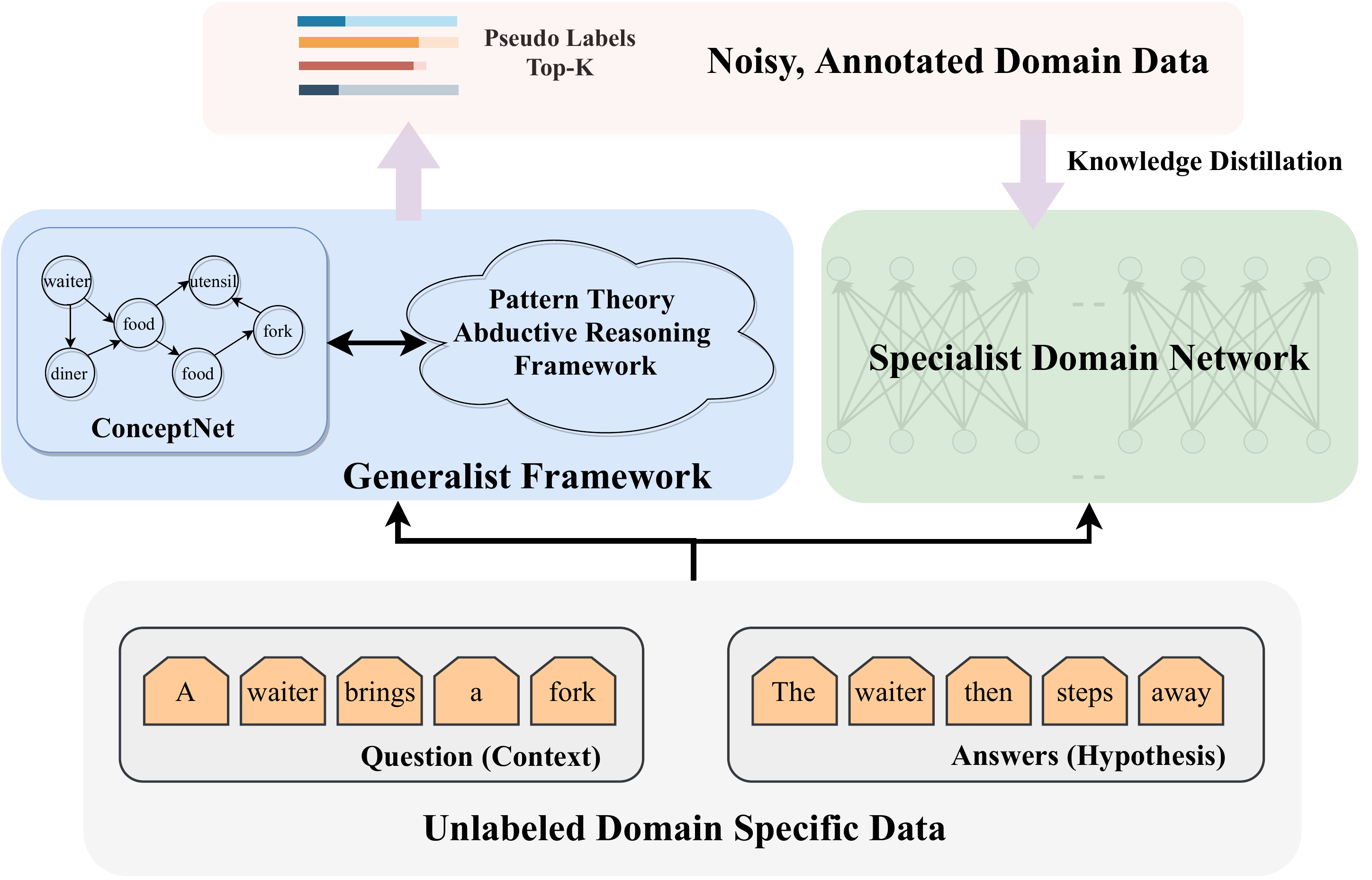}
\caption{\textbf{Proposed Approach}: Given unlabeled, domain-specific data, we run through a generalist, abductive reasoning framework to create noisy, self-labeled data. A domain-specific network is then trained on these labels to build specialists networks. 
} 
\label{fig:overallApproach}
\end{figure}

Motivated by the desire to address the limitation of current, \emph{highly supervised} question answering models and motivated by their success in understanding semantic entailment, we propose a novel self-supervised, abductive reasoning approach that can extract knowledge from large-scale, noisy knowledge bases such as ConceptNet\cite{liu2004conceptnet} and provide supervision for deep learning models to perform commonsense question answering, \emph{without any human-labeled, training data}. In addition to performing open-domain question answering, our model can provide deep contextualized graph representations of the observed evidence for transparent, interpretable decision making. 

We build upon the idea of \textit{abductive reasoning}~\cite{peirce1931collected,peirce1965pragmatism} for knowledge distillation~\cite{buciluǎ2006model,hinton2015distilling} to provide an effective framework for combining prior knowledge in knowledge bases with the representation learning capability of current language models. While abductive reasoning has not been explored to a great extent in existing literature, knowledge distillation has been used to obtain significant improvements in model compression~\cite{chen2017learning} and quantization~\cite{polino2018model}, to name a few. Knowledge Distillation (KD) aims to transfer the \emph{dark knowledge} from large, high performing model(s) to smaller, more compact networks. However, traditional approaches to KD still involve training the teacher models on large, annotated training data. We aim to transfer the general, commonsense knowledge from ConceptNet to domain-specific, training data through abductive reasoning.

To address these limitations, we propose a pattern theory-based abductive reasoning framework, which enables open domain question answering without using training annotations specific to any particular domain. 
A significant departure from current approaches to question answering, we construct a ``\emph{contextualized interpretation}'' of the evidence (the question or context) and each of the provided hypotheses (the answer choices), expressed in a graph-like structure using pattern theory. We define an interpretation to be a connected representation that captures the semantic structure of the evidence. Similar to scene graphs~\cite{xu2017scene} in images, an interpretation is a deeper, meaningful representation of observed concepts (actors, actions and actor-object interactions) as well as unobserved concepts (background knowledge of each concept used to express deeper semantics) or \emph{contextualization cues}. We use such contextualized interpretations to perform ``\emph{inference to the best explanation}'' to find the most plausible hypothesis (answer).

\textbf{Contributions}: We make the following contributions. We 
\begin{itemize}
    \item address the problem of unsupervised commonsense question answering, using no human-annotated training data
    \item introduce the notion of \emph{abductive reasoning} to perform unsupervised commonsense question answering beyond identifying language entailment
    \item show that knowledge distillation can be used to transfer knowledge encoded in large scale, general knowledge bases to train neural networks on domain-specific data
    \item show that model trained using the noisy, self-labeled data can retain up to $75\%$ of the performance of models trained on large amounts of human-annotated training data.
\end{itemize}
    

Further, we evaluate the performance of current, state of the art question answering models under a ``\emph{resource constrained}'' environment. Here, we limit the amount of training data available for five (5) strong baselines and analyze the effect of the reduced training data on their performance. We find that while BERT performs well, even given as few as 100 training examples, the initialization of the weights using the pre-training does not help generalize to out of domain questions. 
\section{Related Work}
\textbf{Question Answering} has been studied to a great extent in literature. Broadly, there are four (4) types of question answering tasks in literature, namely reading comprehension~\cite{dua2019drop,rajpurkar2016squad}, community question answering~\cite{ruckle2019coala}, natural language inference (NLI)~\cite{wang2019superglue,zellers2018swag,zellers2019hellaswag} and visual question answering~\cite{antol2015vqa}. Approaches to each of these question-answering tasks can be classified into two broad categories - semantic similarity matching and relevance matching models. 
Similarity matching models typically involve the computation of semantic similarity between the question and answer representations, typically using a neural network model. The answer with the highest similarity is chosen as the predicted answer choice. Some of the common similarity matching methods include BERT ~\cite{devlin2018bert}, OpenAI GPT ~\cite{radford2019language}, ESIM~\cite{chen2017neural} and LSTM based approaches. Other approaches represent some of the early supervised models such as Bag of Words (BoW) and FastText~\cite{joulin2016bag} models.
The other class of approaches attempts to match answers to the question by quantifying their mutual relevance. The general framework can be described as a \emph{compare, attend, and aggregate} framework~\cite{parikh2016decomposable}. The framework typically begins with a vector representation of the question and answer, computing the relevance of each part and finally aggregating the representation for the final prediction.  


\textbf{Knowledge Distillation} was introduced by Caruana \emph{et al}~\cite{buciluǎ2006model} and generalized by Hinton \emph{et al} ~\cite{hinton2015distilling} as a method to effectively transfer the learned knowledge from larger, more cumbersome models into smaller, more compact networks. It typically involves training the smaller network (the student) with the labels from the larger model (the teacher) presented as soft targets along with the one-hot ground truth annotations. This allows the soft labels (pseudo labels) to act as a regularizer and help the student learn more effective representations. The knowledge distillation framework has been explored for action recognition~\cite{zhang2016real}, quantization~\cite{polino2018model} and model compression~\cite{chen2017learning} to name a few. We extend this idea by eliminating the use of ground truth targets and train exclusively with the pseudo labels as target along with a temperature-based cross-entropy function. 

\textbf{Abductive Reasoning} has not been explored to a great extent in literature, especially from a computational viewpoint. Introduced by Peirce~\cite{peirce1931collected}, abduction refers to ``inference to the most plausible explanation for incomplete observations''. While deemed to be the source of reasoning used by humans in everyday situation~\cite{fischer2001abductive,aliseda2006abductive}, there have been, surprisingly, very few computational models introduced. Many have been logic based abductive reasoning~\cite{elsenbroich2006case,meheus2006formal,singla2011abductive}. Recently Bhagavatula \emph{et al}~\cite{bhagavatula2019abductive} have introduced the task of abductive NLI where the abductive reasoning task is framed as question answering. 
\section{Abductive Reasoning Framework}\label{sec:abductiveReasoning}

Abductive reasoning typically involves the inference to the most \textit{plausible} hypothesis that completes observed evidence. 
This reasoning process typically begins with a set of observations (both complete and incomplete) and attempts to find the most likely explanation for the occurrence of these observation(s). 
At the core of this process is the use of contextual knowledge that allows for evaluating the plausibility of each hypothesis and identifying the hypothesis with maximum evidence to support its validity. 

Formally, we define the abductive reasoning process to be an optimization process that aims to find the optimal hypothesis $H_i \in \{H_1, H_2, H_3, \ldots H_n\}$ to maximize the probability of occurrence conditioned upon the observed evidence $E_t$ and contextual knowledge about the evidence,  $C_t$. This can be expressed as the optimization for 
\begin{equation}
     \argmax_{H_i \in \{H_1, H_2, H_3, \ldots H_n\}} p(H_i | C_t, E_t)
    \label{eqn:AbductiveProb}
\end{equation}
where $E_t$ represents the observed evidence from the input data at time $t$. This optimization involves the empirical computation of the probability of occurrence for each hypothesis $H_i$ given the contextual knowledge $C_t$. 

As opposed to logic-based reasoning, we use natural language to express the data from the evidence and hypotheses. Hence, assigning a likelihood for any given hypothesis requires a complete understanding of the observed evidence, which requires interpreting the semantic structure that links the recognized actors, their actions, and interactions. 
Such understanding involves the modeling of the underlying pattern such as atomicity, regularity, and an inference methodology for using the knowledge of these fundamental properties of the pattern. 

\subsection{Representing Interpretations using Pattern Theory}\label{sec:PT}
\begin{figure}[h]
\centering
\includegraphics[width=0.99\columnwidth]{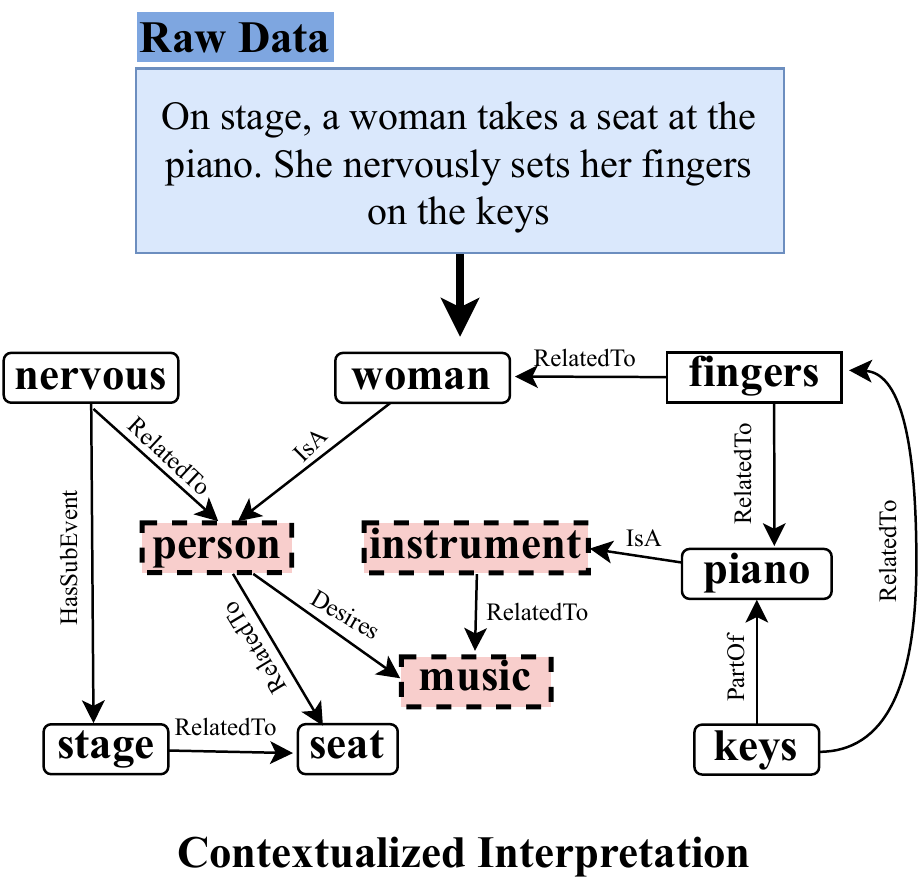}
\caption{An example of how raw data, in the form of natural language sentences, is expressed a contextualized interpretation in the pattern theory framework.
} 
\label{fig:interpretationExample}
\end{figure}

We represent the evidence and hypothesis as an \emph{interpretation} and express it in terms of Grenander's canonical representation of general pattern theory~\cite{grenander1996elements}. 
An example interpretation of a given data is shown in Figure \ref{fig:interpretationExample}. 
Each of the interpretations of the observed evidence is conditioned by the contextual knowledge encoded in large-scale knowledge bases such as ConceptNet\cite{liu2004conceptnet,speer2013conceptnet}. 
The pattern theory formalism allows for a flexible, graphical representation of the observed concepts in the evidence and hypotheses. 
We define \emph{concepts} as both the \emph{observed} attributes of the evidence such as actors, actions, objects, and the actor-object interactions and the \emph{unobserved}, contextual knowledge about these observed concepts. In the example in Figure \ref{fig:interpretationExample}, nodes in white represent the observed concepts, and the nodes in red represent the unobserved concepts. 

\textbf{Concepts as Generators}. In pattern theory, concepts are represented as \textit{generators} $g_i \in G_s$, where $G_s$ is the collection of all generators required to express the semantics of a given environment. 
Each generator $G_i$ represents a single, atomic element that expresses the presence of each concept in the evidence. 
We allow for two different types of generators based on their provenance. 
\emph{Grounded} generators ($\underline{g}_1, \underline{g}_2, \ldots,\underline{g}_q \in G_E$) are concepts whose presence in the interpretation can be grounded to their presence in the evidence. 
\emph{Ungrounded} generators ($\bar{g}_1, \bar{g}_2, \ldots,\bar{g}_q \in G_C$), on the other hand, are concepts that represent the essential, contextual knowledge about grounded generators. 
The term \emph{grounding} is used to differentiate concepts based on their presence in the evidence. 
In Figure \ref{fig:interpretationExample}, the concepts \emph{person, instruments} and \emph{music} are the ungrounded generators, whereas the other concepts represent the grounded generators. 
While the ungrounded generators are not present in the input data, they are essential to understanding the semantic relationship between the actor (\emph{woman}) and the object of interest (\emph{piano}).

\textbf{Expressing Associations using Bonds}. Each of the concepts shares a semantic relationship with other generators. These associations can represent specific semantics such as spatial, temporal, and social, to name a few. We express these semantics in the pattern theory interpretation through links called \emph{bonds}. 
The direction of the \textit{bonds} signifies the semantics of a concept and the type of relationship shared with its bonded generator. For example, the generators \emph{piano} and \emph{instruments} are semantically related through the assertion that ``\textit{a piano is an instrument}''. 
The \emph{energy} of a bond is used to quantify the strength of the semantic relationship expressed between two generators. 
The energy of a bond is given by the function:
\begin{equation}
b_{sem}(\beta^{\prime}(g_i),\beta^{\prime\prime}(g_j)) = \tanh(\phi(g_i,g_j)). 
\label{SemEnergy}
\end{equation}
where $\beta^{\prime}$ and $\beta^{\prime\prime}$ represent the bonds from the generators $g_i$ and $g_j$, respectively; $\phi{.}$ is the strength of the assertion expressed in the bond. Note that we use $\tanh$ to normalize the assertion strength to range from $-1$ to $1$. The normalization range of $-1$ and $1$ also allows us to express negative assertions between two concepts that are not compatible and hence can reduce the validity of the contextualized evidence. We use the labeled assertions from ConceptNet as the source of these bonds, both for quantification as well as labels.

\textbf{Interpretations as Configurations}. The semantics of the given evidence can be expressed through complex structures called \textit{configurations}, $c$. Generators combine through their local bond structures. An example of a configuration is shown in Figure \ref{fig:interpretationExample}. 
Each configuration has an underlying graph topology, specified by a connector graph $\sigma \in \Sigma$, where $\Sigma$ is the set of all available connector graphs. $\Sigma$ is also called the connection type and is allowed to follow the directed connections between elements of a Partially Ordered Set ($POSET$). 
A $POSET$ prescribes a hierarchy for the relationships between the concepts in ConceptNet with ordering present between concepts at different levels of hierarchy. We allow for two levels of hierarchy in the generator space - one for the concepts present in the evidence and one for those in the hypotheses. The evidence generators are one level above those of the hypotheses generators, implying a natural order of connection. 

Formally, we define a configuration $c$ to be a connector graph $\sigma$ whose sites $1, \ldots, n$ are populated by a collection of generators $g_1, \ldots, g_n$ expressed as, 
\begin{equation}
c = \sigma (g_1, \ldots, g_i); g_i \in G_{S}
\label{configEqn}.
\end{equation}
The semantic content of the configuration $c$ is defined by the choice of the generators $g_1, g_2, \ldots g_i$. The configuration in Figure \ref{fig:interpretationExample} can be expressed in terms of the evidence (raw data) and vice versa. 

The probability of a given configuration $c$ can be computed by the energy \emph{\(E(c)\)} of a configuration $c$. The energy is defined to be the sum of the bond energies (Equation \ref{SemEnergy} formed by the bond connections between generators  in the configuration. The energy is given by 
\begin{equation}
\begin{split}
E(c) &= -\sum_{ (\beta^{\prime},\beta^{\prime\prime})\in c}{b_{sem}(\beta^{\prime}(g_i),\beta^{\prime\prime}(g_j))}
\end{split}
\label{energy}
\end{equation}
and the probability of the configuration is given by $P(c) \propto  e^{-E(c)}$. Hence lower the energy of the configuration, the higher its probability. 

\textbf{Knowledge Source: ConceptNet}. To model the semantics of the interpretations, we propose the use of a large commonsense knowledgebase as the source of knowledge about concepts and their semantic associations. 
ConceptNet, proposed by Liu and Singh~\cite{liu2004conceptnet} and expanded to ConceptNet5~\cite{speer2013conceptnet}, is one such knowledge base that maps concepts and their semantic associations into a large scale, traversable semantic network. 
ConceptNet serves as the source of \textbf{general human knowledge} which encodes cross-domain semantic information in a hypergraph. Each node in the ConceptNet framework represents concepts which are connected by weighted edges, labeled as expressed by humans in natural language.

ConceptNet contains more than 3 million concepts, extracted from a variety of sources such as DBPedia, Wiktionary, OpenCyc, and WordNet, to name a few. There are more than 25 assertions (semantic relations) connecting the concepts, with each assertion specifying and quantifying the semantic relationship between the two concepts
The weight of each edge determines the validity of the assertion based on the sources, with positive values indicating positive assertions and negative values indicating the opposite. 
In this work, we consider all the concepts in ConceptNet to be the generator space $G_s$, as well as the source of knowledge for quantifying the bonds between generators. Hence, the weights of the assertions are used to populate the value of $\phi{.}$ in Equation \ref{SemEnergy}, which is also used to determine the validity of the contextualized evidence.

\subsection{Building Contextualized Interpretations}\label{sec:contextualization}
At the core of our approach is the notion of ``\textit{contextualization}''. First defined by Gumperz~\cite{gumperz1992contextualization}, contextualization involves the use of relevant ``\textit{presuppositions}'' from prior knowledge to maintain involvement in the current task. Here, it refers to the use of prior knowledge to aid in interpreting the observed evidence. 
Specifically, ``\textit{presuppositions}'' refers to the inherent knowledge of a concept such as its properties, shared semantics with other concepts and background knowledge of concepts, their properties, and semantics. This use of prior knowledge allows us to go beyond what is observed and construct interpretations beyond simple, pairwise relationships. These presuppositions are also called ``\emph{contextualization cues}'' and represent the ``\textit{ungrounded}'' concept generators in the evidence interpretation.

The use of contextualization cues has two distinct advantages: (1) it allows us to capture semantic relationships among concepts whose co-occurrence has not been observed and (2) it will enable us to move towards an open world paradigm and hence bypass the need for annotated training data. 
Formally, let concepts be represented by $g_i$ for $i=1, \ldots, N$ and let ${_{g_i}R_{g_j}}$ represent relations between two concepts, then contextualization cue, $g_k$, satisfies the following expression $  \mbox{not} \left( {_{g_i}R_{g_j}} \right) \wedge  {_{g_i}R_{g_k}} \wedge {_{g_k}R_{g_j}}$. Hence, two concepts that do have a direct relationship can be correlated using contextualization cues. For example, in Figure \ref{fig:interpretationExample}, the use of contextualization cues \emph{person, music} and \emph{interpretation} allow us to establish a semantic association between the concepts \emph{woman} and \emph{piano}.

Hence, the task of constructing the contextualized evidence then becomes finding an optimal interpretation, $c$, given the evidence generators $E_t$, a set of hypothesis generators $H_i$ and the prior knowledge in terms of the ConceptNet graph, $C_N$. 
We factor this probability into two parts: a likelihood term, $p(G_f|c)$ and a prior, $p (c|C_N)$, normalized by the distribution over the evidence where $G_f$ is the combined set of both evidence and hypothesis generators. Hence constructing the contextualized interpretation then becomes finding the configuration $c$ that maximizes the probability given by
\begin{equation}
p (c | C_N, G_f) = \frac{p(G_f|c) p (c|C_N)}{p(G_f|C_N)}
\label{eqn:contextualizationProb}
\end{equation}
This probability can be captured using energy functions.
\begin{equation}
P(c | C_N, G_f) = \frac{1}{Z} e^{-E(G_f|c) - E(c|C_N)}
\label{probConfigEqn}
\end{equation}
where $E(G_f|c)$ represents the energy of the configuration $c$ that involve the grounded generators and the detected concepts. While, $E(c|C_N)$ captures the energy of the ungrounded, prior, generators. The total energy $E(c)$ of a configuration $c$ is the sum of these energies: $ E(c) = E(G_f|c) + E(c|C_N)$, as defined in Equation \ref{energy}. Each of the terms $E(c|C_N)$ and $E(G_f|c)$ is computed by only summing the energy of all bonds over the ungrounded generators and grounded generators, respectively.
\subsection{IBE: Inference to the Best Explanation}\label{sec:IBE}
\begin{figure*}[h]
\centering
\includegraphics[width=0.90\textwidth]{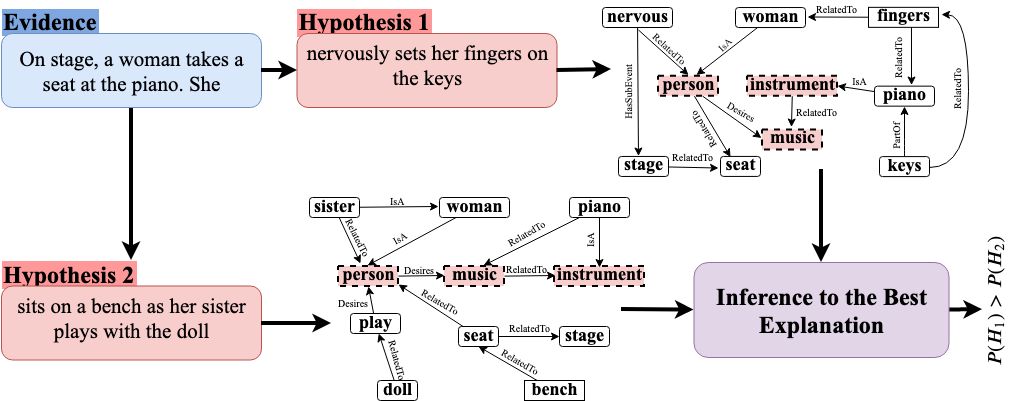}
\caption{The proposed \textbf{Abductive Reasoning Process} is illustrated here. Given an observed evidence and putative hypotheses, contextualized interpretations of the evidence is constructed. A pairwise comparison algorithm is then used to perform \emph{Inference to the Best Explanation} and rank the hypotheses in terms of plausibility.  
} 
\label{fig:abductiveReasoning}
\end{figure*}
The final step in the abductive reasoning framework is \emph{inference to the best explanation}. 
In our framework, this involves the construction of contextualized interpretations for each of the available hypotheses $H_i \in H_n$ along with the observed evidence $E_t$. 
Once the configurations have been constructed, the validity or rather the ``\textit{plausibility}'' of the hypothesis can be obtained by computing the probability of the configuration as defined in Equation \ref{energy}. 
Finding the highest-ranking hypothesis then becomes a matter of \emph{pairwise comparison} between the available set of hypotheses. We use the premise from the Bradley–Terry model~\cite{bradley1952rank} to obtain the outcome of the pairwise comparison between two given configurations, as illustrated in Figure \ref{fig:abductiveReasoning}. 
The pairwise comparison between two contextualized configurations $c_{H_i}$ and $c_{H_j}$ is given by
\begin{equation}
    P(c_{H_i} > c_{H_j}) = \frac{P(c_{H_i})}{P(c_{H_i}) + P(c_{H_j})}
    \label{eqn:pairComp}
\end{equation}
where $P(c_{H_i})$ is the probability of the contextualized interpretation of the evidence $E_t$ and a given hypotheses $H_i$. When performed with all available hypotheses $H_n$, it becomes the optimization for the inference defined in Equation \ref{eqn:AbductiveProb}. Any case of indifference in the outcome of this test is decided by choosing the hypothesis with highest energy among grounded concept generators. 

\subsection{Knowledge Distillation for Domain Specialization}
The knowledge from the abductive reasoning framework is distilled into a specialist neural network by presenting the hypothesis selected from IBE as targets for optimization. The probability of each of the hypotheses produced from the specialist network is given by 
\begin{equation}
    P(H_i) = \frac{exp(q(H_i)/T)}{\sum\limits_{j}^{n} exp(q(H_j)/T)}
    \label{eqn:sftmax_temp}
\end{equation}
where $q(H_i)$ represents the logits layer for the given hypothesis $H_i$ and its corresponding probability is given by $P(H_i)$. $T$ represents the temperature parameter which modulates the probability assigned to each of the target hypotheses. When $T \rightarrow \infty$, all hypotheses have uniform probability and $T = 1$ represents the standard softmax function. 
\section{Experimental Evaluation}
\subsection{Data}
We evaluate the performance of the proposed abductive reasoning approach on two different commonsense NLI datasets in SWAG ~\cite{zellers2018swag} and HellaSWAG~\cite{zellers2019hellaswag}. 
The use of adversarial filtering in both of these datasets ensure that the effect of annotation artifacts is reduced and hence allows us to evaluate the robustness of our approach. 
Additionally, the premise for the construction of both these datasets is the idea predicting which event is most likely to
occur next in a video, given an observation of the current event. This premise offers two significant challenges: (1) answering these questions go beyond what is observed in natural language and requires reasoning across a variety of themes such as social, temporal and spatial to name a few; and (2) the language descriptions are grounded in vision, which makes the reasoning over the language concepts more susceptible to variations in the physical world. We use the official train, dev and test split for both datasets.

The SWAG~\cite{zellers2018swag} dataset consists of 113k multiple choice questions derived from captions of consecutive events of videos in the ActivityNet Captions ~\cite{krishna2017dense} and the Large Scale Movie Description Challenge (LSMDC)~\cite{rohrbach2017movie} datasets. 
The videos cover a variety of domains and hence requires reasoning across domains, temporal scales, and physical interactions to complete the task. 
Each question is accompanied by four (4) answer choices, with one being a human-verified ``gold'' ending and three (3) adversarial distractors.

The HellaSWAG~\cite{zellers2019hellaswag} dataset is a commonsense question-answering dataset consisting of around 70k multiple-choice questions. 
It, like SWAG, is also grounded in vision by constructing the question-answer pairs from the captions of consecutive videos in ActivityNet. 
Additionally, a more challenging domain is introduced by populating question-answer pairs by completing how-to articles from WikiHow, an online how-to manual. 
The task is to choose the most plausible answer choice from a set of four (4) possible answer choices.

\subsection{Evaluation Metrics and Baselines}
We use several fully supervised and weakly supervised baselines to evaluate our approach. We ensure that we have a balanced mix of neural network-based approaches as well as classic approaches. The fully supervised baselines include GPT~\cite{radford2019language}, BERT~\cite{devlin2018bert}, fastText~\cite{joulin2016bag}, ESIM ~\cite{chen2017neural} and an LSTM-based approach. 
For comparison with weakly supervised approaches, we evaluate with models trained on the SNLI task for producing to obtain a probability set for entailment, neutral, and contradiction. A bilinear model is trained only to convert SNLI probabilities to answer probabilities. We essentially learn only the correlations between the entailment categories of SNLI to the probability space in the QA task. 
We evaluate all approaches by computing the accuracy of the predictions. We use the official scoring protocols provided by the authors of SWAG and HellaSWAG for a fair comparison with the other methods. 

\subsection{Quantitative Evaluation}
We first evaluate our approach on the \textbf{SWAG} dataset and compare against fully supervised and weakly supervised. The results are presented in Table \ref{table:perfSWAG}. We show the performance of fastText to highlight the importance of commonsense knowledge and the abductive reasoning process. FastText models a given text as a bag of n-grams and predicts the probability of each ending being correct or incorrect. This approach is heavily reliant on word embeddings and does not generalize well to this task. It is also interesting to note that our approach outperforms all weakly supervised baselines such as the dual bag of words approach and the SNLI-based approaches. These approaches are the closest related to our approach since they are not trained directly on the SWAG training split. 
\begin{table}[h]
\centering
\begin{tabular}{|c|c|c|c|}
\hline
\multirow{2}{*}{\textbf{Supervision}} & \multirow{2}{*}{\textbf{Approach}}      & \textbf{Val.}  & \textbf{Test}\\ 
  &    & \textbf{Acc.}  & \textbf{Acc.}\\ \hline
\multirow{6}{*}{Full} 
 & fastText & 29.4 & 28.0 \\
 & LSTM + GloVe & 43.1 & 43.6\\
 & DecompAttn. + GloVe & 47.4 & 47.6 \\
 & ESIM + GloVe & 51.9 & 52.7\\
 & ESIM + ELMO & 59.1 & 59.2\\
 & OpenAI GPT & 77.6 & 77.9\\
 & BERT & \textbf{86.6} & \textbf{86.3}\\ 
 \hline
\multirow{2}{*}{Weak}
 & DualBoW+GloVe & 34.5 & 34.7 \\
  & SNLI + DecompAttn.  & 35.8 & 35.8 \\ 
  & SNLI + ESIM & \textbf{36.4} & \textbf{36.1} \\
  \hline
\multirow{2}{*}{None} & Ours PT (No Training) & 38.4 & 38.2\\
 & PT +  BERT & \textbf{39.7} & \textbf{39.5}\\

\hline
\end{tabular}
\caption{Performance on SWAG data set}
\label{table:perfSWAG}
\end{table}
BERT currently has the best performance on the validation and test sets on the SWAG test set. It should be noted that the fully supervised approaches required significantly more training data - both in the form of labels as well as training epochs. We can successfully transfer the knowledge onto the BERT architecture using our abductive reasoning approach with just $1$ epoch of training while retaining $46\%$ of the original model's performance without any human annotations. 

We also evaluate our approach on the tougher \textbf{HellaSWAG} dataset. The adversarial filtering technique on this dataset has been improved to increase the perplexity of BERT-like models on this task. The effect of this approach can be seen in Table \ref{table:perfhSWAG}. Again, we compare against the same baselines as in SWAG and find that our approach offers competitive performance to the fully supervised baselines. We find that using our abductive reasoning approach on BERT allows the model to retain up to $59\%$ of its performance as compared to a model trained directly on the annotations. 

\begin{table}[h]
\centering
\begin{tabular}{|c|c|c|c|}
\hline
\multirow{2}{*}{\textbf{Supervision}} & \multirow{2}{*}{\textbf{Approach}}      & \textbf{Val.}  & \textbf{Test}\\ 
  &    & \textbf{Acc.}  & \textbf{Acc.}\\ \hline
\multirow{6}{*}{Full} & FastText & 30.9 & 31.6\\
& ESIM + ELMO & 33.6 & 33.3\\
 & LSTM + GloVe & 31.9 & 31.7\\
 & OpenAI GPT & 41.9 & 41.7\\
 & BERT & \textbf{46.7} & \textbf{47.3}\\ \hline
\multirow{2}{*}{None} 
 & PT +  BERT & 27.8 & 28.1\\
 & Ours PT (No Training) & \textbf{28.3} & \textbf{26.7}\\

\hline
\end{tabular}
\caption{Performance on HellaSWAG data set}
\label{table:perfhSWAG}
\end{table}

It should be noted that the use of ConceptNet-based contextualization on HellaSWAG has one significant drawback: the answers are too similar in their use of semantics, and hence the potential for indifference goes up. We observe that the number of examples where the second-best hypothesis had the same energy as the top hypothesis was $35\%$ when the correct answer was in the top 2. This indifference could, arguably, be diminished by grounding the concepts in vision or other modalities. This is not an unreasonable assumption since the pre-training data used in BERT contains articles from Wikipedia which have vision-based explanations which constrain the semantics to those observed in vision.

\begin{figure}[h]
\centering
\includegraphics[width=1.0\columnwidth]{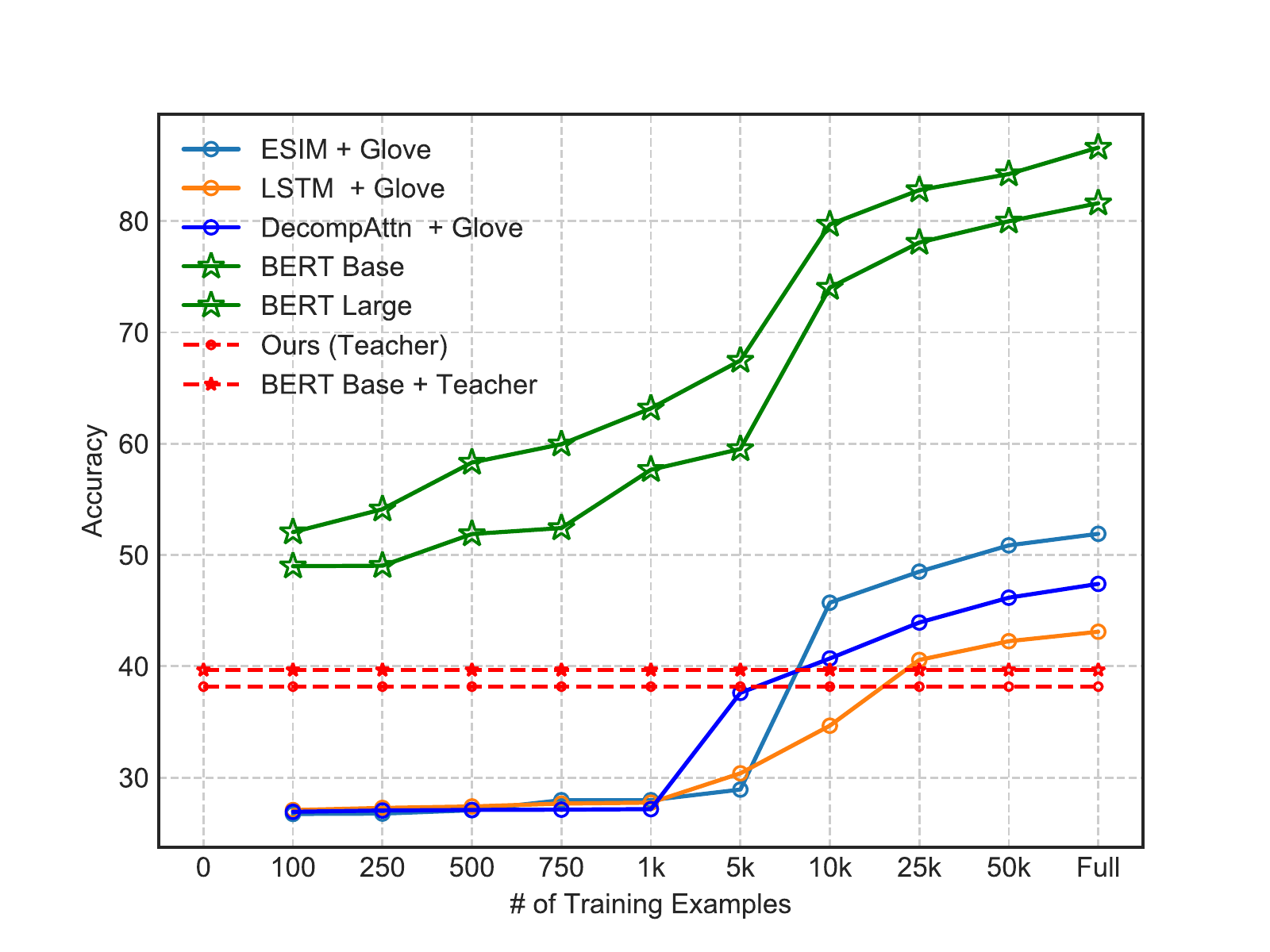}
\caption{Comparison of current QA models' performance as a function of number of available training questions under a low resource setting on the SWAG dataset.} 
\label{fig:dataPlot}
\end{figure}

We evaluate the performance of current QA systems under a \textbf{Low Resource} environment. We limit the amount of training data available to these models for a fair comparison with our approach. The results are shown in Figure \ref{fig:dataPlot}. We plot the accuracy of the approaches on the SWAG validation data vs. the number of training samples available to them. We present the average result over $5$ runs with randomly sampled examples from the training data. We can see that most of the current approaches perform well when presented with increasingly large amounts of data. BERT seemingly performs well when presented with as few as $100$ training samples, achieving $52.1\%$ on the validation set. %
However, given their ability to be rapid surface learners (CITE HellaSWAG), we evaluate the ability to \textbf{generalize to new domains} by evaluating the models on the HellaSWAG validation data while training on the SWAg data. As can be seen from Table \ref{table:genVocab}, there is a significant gap in generalization ability of BERT from SWAG to HellaSWAG, especially in a low resource constraint. We need at least 10,000 training examples on SWAG for it to outperform our model on HellaSWAG, although both datasets are derived from the same source. 

\begin{table}[h]
\centering
\begin{tabular}{|c|c|c|}
\hline
\textbf{Number of Samples} & \textbf{Approach}     & \textbf{Accuracy} \\ \hline
\multirow{2}{*}{100} & BERT & 23.4\\
 & ESIM + ELMO & 19.6 \\ \hline
\multirow{2}{*}{1,000} & BERT & 26.1\\
 & ESIM + ELMO & 20.3 \\\hline
 \multirow{2}{*}{5,000} & BERT & 29.0\\
 & ESIM + ELMO & 20.7\\
\hline
\multirow{2}{*}{10,000} & BERT & 30.7\\
 & ESIM + ELMO & 21.5\\
\hline
\multirow{2}{*}{25,000} & BERT & 32.8\\
 & ESIM + ELMO & 23.9\\
\hline
\multirow{2}{*}{All (73,000)} & BERT & 34.6\\
 & ESIM + ELMO & 26.7\\\hline
None & Our Approach & \textbf{28.3}\\
\hline
\end{tabular}
\caption{Generalization from SWAG to new domains and vocabulary introduced in HellaSWAG}
\label{table:genVocab}
\end{table}

We also perform \textbf{ablative studies} to evaluate our approach by varying the student model and the source of semantic knowledge. We train two student networks other than BERT. We use two strong models as student networks, namely ESIM and Unary LSTM models, which achieve $51.9\%$ and $43.1\%$ respectively on SWAG. 
We also vary the source of semantic knowledge by using GloVe~\cite{pennington2014glove} representations and ConceptNet NumberBatch~\cite{speer2017conceptnet} and using dot-product to compute the likelihood of co-occurrence of two concepts. We also test the effectiveness of using contextualized interpretations by relying only on the direct semantic relationships between two concepts in the ConceptNet semantic network. As can be seen from Table \ref{table:ablative}, the use of ConceptNet is essential for robust abductive reasoning. 
\begin{table}[h]
\centering
\begin{tabular}{|c|c|}
\hline
\multirow{2}{*}{\textbf{Approach}}     & \textbf{Val.} \\
 & \textbf{Acc.}\\\hline
Ours PT (NB only)  & 25.9 \\
Ours PT (GloVe only)  & 26.3 \\
Ours PT (GloVe + NB)  & 28.1 \\
Ours PT (No Contextualization)  & 33.6 \\
Ours PT (Full Model) & 38.4 \\
\hline
PT + LSTM+GloVe & 32.4\\
PT + ESIM+GloVe & 39.4\\
PT +  BERT & \textbf{39.7}\\
\hline
\end{tabular}
\caption{Ablative Studies for our model, where we compare against variations of our approach. We compare different source of knowledge and different student networks.}
\label{table:ablative}
\end{table}
Using representations learned from pre-computed embeddings such as GloVe or Numberbatch without using ConceptNet do not generalize to the QA with adversarial filtering. Additionally, the use of contextualization to construct interpretations also has a positive effect on the robustness of our approach. We can see improvements of $6.3\%$ absolute percentage points in accuracy through the use of contextualization. We do not see any increase in performance by combining all three sources of knowledge. 
\section{Discussion and Future Work}
We present one of the first works on abductive reasoning for commonsense question answering. We show that the use of a global source of knowledge can be used to distill commonsense knowledge and reasoning to neural networks without large amounts of annotations. We demonstrated the use of pattern theory to express semantics in the evidence in a highly interpretable, contextualized interpretation for validating the plausibility of natural language expressions without training highly expensive models. Extensive experiments demonstrate the applicability of the approach to different domains and its highly competitive performance. We aim to ground the contextualized interpretations in vision to enable automatic discovery of concepts without the need for relearning or large amounts of training data.

{\small
\bibliographystyle{aaai}
\bibliography{egbib}
}

\end{document}